\documentclass{article}

\usepackage[final, nonatbib]{nips_2016} 

\usepackage[utf8]{inputenc} 
\usepackage[T1]{fontenc}    
\usepackage{url}            
\usepackage{booktabs}       
\usepackage{amsfonts}       
\usepackage{nicefrac}       
\usepackage{microtype}      

\usepackage{amssymb}
\usepackage{amsthm}
\usepackage{xspace}
\usepackage{amsmath}
\usepackage{setspace}
\usepackage{amscd}
\usepackage[numbers]{natbib}
\usepackage{subfigure}
\usepackage{tikz}
  \usepackage{algorithm}
\usepackage{algorithmic}
\usepackage{dsfont}
\usepackage{comment}

\newcommand{\x}{\mathbf{X}}
\newcommand{\z}{\mathbf{Z}}
\newcommand{\y}{\mathbf{Y}}
\newcommand{\qq}{\mathbf{q}}
\newcommand{\rr}{\mathbf{r}}
\title{Discriminative Neural Topic Models}

\author{
  Gaurav Pandey and Ambedkar Dukkipati\\
  Department of Computer Science and Automation\\
  Indian Institute of Science\\
  Bangalore, India \\
  \texttt{\{gaurav.pandey, ad\}@csa.iisc.ernet.in} \\
}

\begin{document}

\maketitle

\begin{abstract}
We propose a neural network based approach for learning topics from text and image datasets. The model makes no assumptions about the conditional distribution of the observed features given the latent topics. This allows us to perform topic modelling efficiently using sentences of documents and patches of images as observed features, rather than limiting ourselves to words. Moreover, the proposed approach is online, and hence can be used for streaming data. Furthermore, since the approach utilizes neural networks, it can be implemented on GPU with ease, and hence it is very scalable. 
\end{abstract}

\section{Introduction} \label{Introduction} 
Mixed membership modeling is the task of assigning the observed features of an object to latent classes. Each object is assumed to be a mixture over the latent classes, and these classes are shared among the objects. This is distinct from mixture modeling, where all the features of the object are assumed to be sampled from a single latent class.

Traditional approaches to mixed membership modeling assume a parametric form for the distribution of a latent class over the features, and the distribution of an object over these classes. For instance, in case of documents, the observed features are the words, and the latent classes are the topics. A topic is assumed to have a multinomial distribution over the words, while a document is assumed to have a multinomial distribution over the topics~\citep{hofmann1999probabilistic}. This statistical model is also referred to as probabilistic latent semantic indexing (pLSI). One can further define Dirichlet priors on the parameters of the multinomial distributions, to obtain a full Bayesian treatment of topic modeling~\citep{blei2003latent, pritchard2000inference}. This model, commonly referred to as latent Dirichlet allocation~(LDA), is widely used for finding topics in document collections, and inferring population structures from genotype data.

Both LDA and pLSI are generative models, whereby the conditional distribution of the observed features given the latent classes is explicitly modeled. This requires explicit knowledge about the parametric form of these distributions. The number of unique words in a language are finite, and hence, it is possible to make these distributions as general as possible by choosing them to be multinomial. However, as the set of all unique observed features becomes comparable to the size of the collection, this approach becomes less and less meaningful. 

Hence, while generative topic models are useful for modeling words in a document collection, they can't be directly used for modeling sentences and paragraphs in documents, or pixels of an image. An intermediate preprocessing step is required that transforms all the observed features in the collection to a relatively small set of unique `codewords'~\citep{fei2005bayesian}. Since this intermediate step is independent of topic modeling, it can result in suboptimal features. Finding good intermediate representations suitable for topic modeling can be as challenging as finding the topics themselves.

While the observed features can be extremely complicated to model (for instance, the pixels in an image), the conditional distribution of the latent classes given the observed features is always multinomial. The parameters of the multinomial distribution are functions of the observed features. In this paper, we propose a model for learning the latent classes of an object without explicitly modeling the observed features. This allows us to make minimal assumptions about the distributions of the observed features. In particular, we propose discriminative neural topic models (DNTM), whereby the conditional distribution of the latent classes or topics given the observed features is directly modeled using neural networks. The model can be trained end-to-end using backpropagation. To our knowledge, this is the first approach to topic modeling that doesn't explicitly model the distribution of observed features given the latent classes.

In order to establish that DNTM indeed learns meaningful topics, we use it for assigning words to topics for well-known text corpuses. The learnt topics are used in secondary tasks such as clustering. Despite the fact that LDA makes minimal assumptions for modeling words in document collections, our model is able to outperform LDA even for these datasets. Note that these experiments only serve as a sanity check confirming that the proposed model indeed learns meaningful topics, and is competitive with LDA even for tasks where LDA makes minimal assumptions.

We also use DNTM for learning topics on CIFAR-10 dataset in an unsupervised manner. We use a convolutional neural network to convert the image into spatial features, which correspond to the words of the document. The CNN is trained by the backpropagating the gradient from the topic model, that is, we train the CNN to learn features, that correspond to meaningful topics. In order to prevent the model from overfitting, we use adversarial training by coupling it with a generator~\citep{goodfellow2014generative}. The model is trained to perform poorly on generated images, while the generator is trained to generate images that perform well on the topic modelling task.

\section{Discriminative Neural Topic Models}
Let $\x^{(1)}, \ldots \x^{(n)}$ be the observed features for $n$ objects, while $\z^{(1)}, \ldots, \z^{(n)}$ be the corresponding latent classes. Let $K$ be the number of latent classes.
\subsection{Modeling the words in a document}
In case of documents, $X^{(i)}_j$ represents the embedding vector of the $j^{th}$ word in the $i^{th}$ document $\x^{(i)}$ with an arbitrary order, and $Z_j^{(i)}$ represents the corresponding topic or latent class. In particular, there exists an embedding vector for every word in the vocabulary. The embeddings are initialized randomly and are trained by backpropagating the gradient of the topic model. 

The length of the $i^{th}$ document is given by $m_i$. We assume that given the words of the document, the topics are independently distributed, that is,
\begin{equation}
P_{\theta}(\z^{(i)}|\x^{(i)}) = \prod_{j=1}^{m_i} P_{\theta}(Z_j^{(i)}|\x^{(i)})\, ,
\end{equation}
$Z_j^{(i)}, 1 \le j \le m_i$ are multinomial random variables whose parameters are functions of $\x^{(i)}$.  An explicit parametric form for this distribution is specified in the experiments section. In the rest of the paper, the dependence of the distributions on $\theta$ is assumed without being explicitly stated.

Since, we wish to associate each word $X_j^{(i)}$, with a single topic $Z_j^{(i)}$ with high confidence, we minimize the entropy of $P(Z_j^{(i)}| \x^{(i)}),\, 1\le j \le m_i$, that is
\begin{equation}
\mathcal{H}(Z_j^{(i)}|\x^{(i)}) = -\sum_{k=1}^K P(Z_j^{(i)} = k|\x^{(i)}) \log P(Z_j^{(i)} = k|\x^{(i)})
\end{equation}
Minimizing entropy ensures that the conditional distribution over the topics for a word is highly peaked for a very few topics. This is also referred to as cluster assumption, and has been used for clustering~\citep{krause2010discriminative} and semi supervised learning~\citep{grandvalet2004semi, chapelle2005semi}, where it enjoys considerable success.  

Next, we need to ensure that the words in a single document have similar distribution over the topics. This corresponds to minimizing the variance between the probability distributions over topics for the words of the document. We achieve this by minimizing the KL-divergence between the conditional distribution of topics given the words $P(Z_j^{(i)}|\x^{(i)})$, and the corresponding average over the words, that is
\begin{equation}
KL(P_{Z_j^{(i)}}||\bar{P}_{\z^{(i)}}; \x^{(i)}) = \sum_{k=1}^K P(Z_j^{(i)} = k|\x^{(i)}) \log \frac{P(Z_j^{(i)} = k|\x^{(i)})}{P_d(k|\x^{(i)})} \,
\end{equation}
where $P_d(k|\x^{(i)})$ is the distribution over the topics for document $d$. If we assume that each observed word in the document occurs with equal probability, then the document distribution over the topics is given by  $P_d(k|\x^{(i)}) = \frac{1}{m_i} \sum_{j=1}^{m_i}P(Z_j^{(i)}=k|\x^{(i)})$.

Finally, in order to ensure that all the documents don't get assigned the same topic, we add a term for encouraging balance among the topics. Towards this end, we define
\begin{equation}
\bar{P}(k) = \frac{1}{n}\sum_{i=1}^n P_d(k|\x^{(i)})
\end{equation}
We maximize the entropy of the above distribution to enforce a uniform prior on the topics. Ways to relax this assumption are discussed later in the paper.

Combining the above three criteria, the objective function for the $j^{th}$ word of $i^{th}$ document is given by
\begin{equation} ~\label{eq:topicModelling}
F_{ij}(\theta) = \mathcal{H}(Z_j^{(i)} | \x^{(i)}) + KL(P_{Z_j^{(i)}}||{P_d}_{\z^{(i)}}; \x^{(i)}) - \mathcal{H}(\bar{P})\,,
\end{equation}
where $\theta$ is the parameter of the distribution $P(Z_i|\x^{(i)})$.
The above term is minimized for all the words of all the documents in the corpus, that is, $\sum_{i=1}^n\sum_{j=1}^{m_i} F_{ij}(\theta)$.

\subsection{Regularizing the model}
In order to ensure that the model doesn't overfit the training data, we use negative sampling. In particular, for text documments, we randomly sample the words in the vocabulary, to create a fake document. Next, we force the model to perform poorly on the fake document as follows:
\begin{enumerate}
\item The distributions over topics for the words in a fake document should be highly uncertain. This corresponds to maximizing the entropy of the corresponding topic distributions, that is, $\mathcal{H}(Z_j^{(i)}|\x^{(i)})$ is maximized for $1\le i\le n, 1\le j \le m_i$.
\item The distribution over topics for the words in a fake document, should have high variance. This corresponds to maximizing the KL-divergence of the topic distribution from the average for the words in the fake document, that is $KL(P_{Z_j^{(i)}}||{P_d}_{\z^{(i)}}; \x^{(i)})$ is maximized.
\end{enumerate}

\subsection{Document Clustering}
Firstly, we verify that the model indeed performs topic modelling. Towards that end, we apply our model to learn topics on 20 newsgroup, one of the most widely used datasets for text categorization. In order to ensure reproducibility, we use a pre-processed version of the dataset~\footnote{The pre-processed dataset is available at https://sites.google.com/site/renatocorrea02/textcategorizationdatasets}. The preprocessed 20 newsgroup dataset consists of 11,293 documents for training and 7,528 documents for testing, with a vocabulary size of 8,165 stemmed words. The data is almost evenly divided between the 20 classes. We discard documents with less than 2 words. 

For clustering tasks, we combine the training and test sets. We use two metrics to evaluate the effectiveness of the proposed topic model for clustering - purity and normalized mutual information (NMI)~\footnote{Details about the two metrics can be obtained at http://nlp.stanford.edu/IR-book/html/htmledition/evaluation-of-clustering-1.html}.  

We compare the proposed model against several standard algorithms for clustering and topic-modelling. These include K-means, Normalized cuts~\citep{shi2000normalized}, probabilistic Latent Semantic indexing (pLSI) and Latent Dirichlet Allocation (LDA). For LDA, pLSI and the proposed model DNTM, the topics correspond to clusters and a document is assigned to the cluster/topic with the highest posterior probability for the given document. This approach has been shown to be more effective than clustering the topic representations of documents~\citep{lu2011investigating}.

The clustering results for 20 newsgroup are given in Table~\ref{20ngclustering}. As can be seen from the results, there is a distinct improvement in performance as one moves from standard clustering algorithms to topic-modelling algorithms such as pLSI, LDA and DNTM. Moreover, the performance obtained using DNTM is at par with the performance using LDA and pLSI.

\begin{table} \label{ss}
\caption{Clustering results on text datasets}
\centering
\subfigure[20 newsgroup dataset]{
\label{20ngclustering}
\begin{tabular}{|c|c|c|}
\hline
Method & Purity & NMI \\ 
\hline
K-means & 34.3\% & 32.4\% \\
Normalized Cut & 23.1\% & 21.7\% \\
pLSI & \textbf{57.7}\% & 56.5\% \\
LDA & 54.7\% & 55.3\% \\
DNTM & 56.5\% & \textbf{56.7}\% \\
\hline
\end{tabular}
}

\end{table}

\subsection{The topics}
Although the proposed model DNTM only learns the distribution of the topics given the words, we can obtain the distribution of the words given the topics as follows: First, we compute the joint probability of the topic $t$ and the word $w$ occurring together as follows:
\begin{align*}
\bar{P}(t,w) = \frac{1}{n}\sum_{i=1}^n \frac{1}{m_i}\sum_{j=1}^{m_i} P(Z^{(i)}_j=t| X^{(i)}_j=w) P(X_j^{(i)}=w)
\end{align*}
Here, $P(X_j^{(i)}=w)=1$, if the $i^{th}$ word of the $j^{th}$ document is $w$ and $0$ otherwise. Informally, the above equation simply computes the probability of observing the topic $t$, every time the word $w$ occurs, and takes the average of all these probabilities. Finally, to compute the posterior, we normalize the above distribution.
\begin{equation}
\bar{P}(w|t) = \frac{\bar{P}(w,t)}{\sum_{{w'} \in \text{vocabulary}} \bar{P}({w'}, t)}
\end{equation}
Using the above method, we obtain the distribution of the topics over the words for DNTM. The $20$ most probable words for each topic obtained using DNTM are listed in Table~\ref{DNTMTopics}. One can observe that the topics learnt by the model are coherent, that is, the words corresponding to a single topic occur commonly in a single class.

\begin{table} \label{DNTMTopics}
\caption{Top 10 stemmed words from 10 randomly selected topics learnt on the 20 newsgroup dataset by DNTM. Note that each topic is coherent,  that is, the words corresponds to a specific class of 20 newsgroup.} 

\begin{tabular}{|c|l|}
\hline
Topic 1 & 'line' 'power'	'sound'	'tape'	'radio'	'cabl'	'switch'	'light'	'phone'	'work' \\
\hline
Topic 2 & 'effect'	'drug'	'medic'	'doctor'	'food'	'articl'	'health'	'research'	'school' \\
\hline
Topic 3 & 'drive'	'card'	'mac'	'driver'	'problem'	'disk'	'monitor'	'appl'	'video'	'work' \\
\hline
Topic 4 &'game'	'team'	'plai'	'fan'	'player'	'win'	'hockei'	'score'	'season'	'playoff'\\
\hline
Topic 5 &'kill'	'world'	'war'	'muslim'	'death'	'armenian'	'attack'	'peopl'	'histori'	'citi'\\
\hline
Topic 6 &'car'	'engin'	'bike'	'dod'	'ride'	'articl'	'road'	'bmw'	'mile'	'front'	\\
\hline
Topic 7 &'state'	'govern'	'israel'	'american'	'isra'	'right'	'nation'	'clinton'	'presid'	'polit'\\
\hline
Topic 8 & 'file'	'program'	'softwar'	'graphic'	'imag'	'color'	'code'	'version'	'format'	'ftp' \\
\hline
Topic 9 & 'sale'	'price'	'bui'	'sell'	'offer'	'interest'	'compani'	'book'	'ship'	'cost' \\      
\hline
Topic 10 &  'god'	'christian'	'jesu'	'love'	'church'	'bibl'	'faith'	'sin'	'christ'	'word'\\
\hline
\end{tabular}
\vspace{.5cm}
\end{table}

\section{Topic modelling in images}
In order to extend the proposed model for images, one needs to define `words' and `documents' for images.
The most common approach for obtaining words from images involves extraction of features (SIFT, HOG etc.,) from images. These features are then clustered using K-means. The corresponding quantized features are used as words~\cite{li2009towards, chong2009simultaneous, shi2013bayesian} for topic modelling.  Learning of features happens independently of topic modelling, thereby resulting in suboptimal features for topic modelling. 
Another approach that has been considered in~\cite{torralba2011learning}, involves training a deep Boltzmann machine on the pixels of an image. The samples from the deepest latent layer of DBM are then used as words for training a hierarchical Bayesian model. 

In this work, however, we learn the words from the pixels of an image by employing a convolutional network.
In particular, let $\x^{(i)}$ represents the $i^{th}$ image, $Y^{(i)}_j$ represents the $j^{th}$ word of the $i^{th}$ image and $Z^{(i)}_j$ represents the corresponding topic. In order to convert the the $\x^{(i)}$ into words $Y^{(i)}_j$, we feed $\x^{(i)}$ to a convolutional neural network. The output of the CNN is treated as words of the image. For instance, if the output of CNN consists of 100 features maps of size 8x8, we treat them as 64 words, where each word is represented using 100 dimensions. The topic distribution of a word is obtained by applying $1\times 1$ convolution to the word representation, and then applying softmax to the output. Note that the words $Y^{(i)}_j$  are deterministic functions of $\x^{(i)}$, and hence, allow the backpropagation of gradient from $Y^{(i)}_j$
to the parameters of the CNN.

Now that we have defined the words in an image, we need to define the concept of a document. The definition of a document depends on the underlying task. In this paper, we consider two possible definitions for documents. In the first scenario, we are interested in clustering similar images together. For this task,
we consider that all the words in an image belong to the same document. This is the usual assumption in topic modelling. The words and the document are then fed to the proposed model, exactly as for documents.

\begin{figure} \label{imageTopics}
\centering     
\subfigure[Topic 1]{\label{fig:a}\includegraphics[width=40mm]{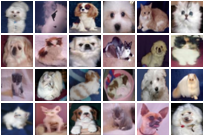}}
\subfigure[Topic 2]{\label{fig:b}\includegraphics[width=40mm]{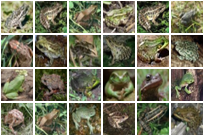}}
\subfigure[Topic 3]{\label{fig:a}\includegraphics[width=40mm]{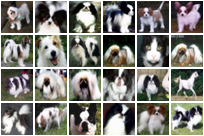}}
\subfigure[Topic 4]{\label{fig:b}\includegraphics[width=40mm]{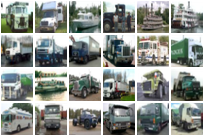}}
\subfigure[Topic 5]{\label{fig:a}\includegraphics[width=40mm]{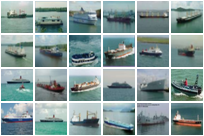}}
\subfigure[Topic 6]{\label{fig:b}\includegraphics[width=40mm]{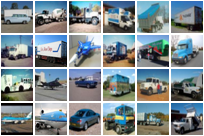}}
\subfigure[Topic 7]{\label{fig:b}\includegraphics[width=40mm]{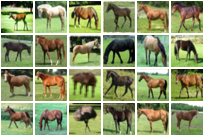}}
\subfigure[Topic 8]{\label{fig:a}\includegraphics[width=40mm]{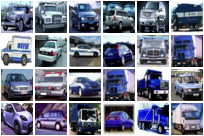}}
\subfigure[Topic 9]{\label{fig:b}\includegraphics[width=40mm]{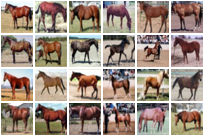}}

\caption{Some of the topics learnt by DNTM on CIFAR-10. For each topic, the 24 images with the highest probability for the given topic are listed  in the figure.}
\end{figure}

\subsection{Treating images as documents}
We use the proposed model to learn topics in CIFAR-10 dataset. The dataset consists of 50,000 training images and 10,000 test images divided into $10$ classes. This dataset is quite challenging, since there is high variability within each class, even though the individual images are only 32x32 pixels. 

We use convolutional neural networks for obtaining the features from the images. In particular, for the 32x32 CIFAR-10 images, we apply 4 layers of strided convolution and ReLU nonlinearity to obtain 64 (8x8) features per image. These features function as words, and are fed as input to the DNTM. Note that CNN is trained only by the DNTM, thereby coupling feature extraction and topic modelling.

We train the DNTM to extract $100$ topics from the CIFAR-10 dataset. $9$ of those topics are shown in Figure~\ref{imageTopics}. In particular, for each topic, we have listed the $24$ most probable images. One can observe that the topics are qualitatively coherent.

We evaluate the purity of the learned topics by retrieving the $100$ most probable CIFAR-10 images for each  topic. The most common label among the $100$ images associated with the topic, is then assigned to the topic. The purity of a given topic is the fraction of the retrieved images that have the same label as the one assigned to the topic. Using the proposed model, we obtain a mean purity of \textbf{49.3}\% for the learned topics.

\bibliographystyle{unsrt}
\bibliography{deep}

\end{document}